\documentclass[letterpaper, 10 pt, conference]{ieeeconf}  

\usepackage{amsmath} % assumes amsmath package installed
\usepackage{amssymb}  % assumes amsmath package installed
\usepackage{amsfonts} % assumes amsfonts package installed
\usepackage{dsfont}

\usepackage[utf8]{inputenc}

\usepackage[pdftex]{graphicx}
\usepackage[usenames, dvipsnames]{color}
\usepackage{color,soul}
\setstcolor{red}

\usepackage[labelformat=simple]{subcaption}

\usepackage{multirow}
\usepackage{makecell}
\usepackage[linesnumbered,ruled]{algorithm2e}

\usepackage{array}
\usepackage{url}

\usepackage{nomencl}

\usepackage{amsthm}
\usepackage{tikz}

\theoremstyle{definition}

\theoremstyle{remark}

\usepackage{makecell}
\usepackage{amssymb}

\usepackage{lipsum}% http://ctan.org/pkg/lipsum
\usepackage{multicol}% http://ctan.org/pkg/multicols

\usepackage{bm}

\usepackage{relsize}

\usepackage{epsfig}
\usepackage{graphicx}
\usepackage{array}
\usepackage{booktabs}
\usepackage{xcolor}
\usepackage{cite}
\usepackage{multirow}

\usepackage{hyperref}
\usepackage{cleveref}

\hypersetup{
    colorlinks=true,
    linkcolor=blue,
    filecolor=magenta,      
    urlcolor=magenta,
}

\usepackage{footnote}
\makesavenoteenv{tabular}
\makesavenoteenv{table}

\newcommand{\xhdr}[1]{\vspace{5pt} \noindent {\textbf{#1} }}
\newcommand{\etal}{\textit{et al}.}

\IEEEoverridecommandlockouts 
\definecolor{mypurple}{rgb}{0.851,0.823,0.908}
\definecolor{mygreen}{rgb}{0.812,0.878,0.823}
\definecolor{mypink}{rgb}{0.914,0.824,0.863}
\definecolor{myyellow}{rgb}{0.930,0.851,0.769}
\definecolor{mygray}{rgb}{0.745,0.745,0.745}

\title{\LARGE \bf
Unsupervised Traffic Accident Detection in First-Person Videos
}

\author{Yu Yao$^{1*}$, Mingze Xu$^{2*}$, Yuchen Wang$^{2}$, David J. Crandall$^{2}$, Ella M. Atkins$^{1}$%
\thanks{$^{*}$The first two authors contributed equally.}%
\thanks{$^{1}$Robotics Institute, University of Michigan, Ann Arbor, MI 48109, USA.
{\tt\footnotesize \{brianyao,ematkins\}@umich.edu}}%
\thanks{$^{2}$School of Informatics, Computing, and Engineering, Indiana University, Bloomington, IN 47408, USA.
{\tt\footnotesize \{mx6,wang617,djcran\}@iu.edu}}%
}
\begin{document}
\maketitle
\thispagestyle{empty}
\pagestyle{empty}

\begin{abstract}
    Recognizing abnormal events such as traffic violations and
    accidents in natural driving scenes is essential for successful
    autonomous driving and advanced driver assistance systems. However, most
    work on video anomaly detection suffers from two crucial
    drawbacks.  First, they assume cameras are fixed and videos have
    static backgrounds, which is reasonable for surveillance
    applications but not for vehicle-mounted cameras.
    Second, they pose the problem as one-class classification, 
    relying on arduously hand-labeled training datasets that limit
    recognition to anomaly categories that have been explicitly trained.
    This paper proposes an unsupervised approach for traffic
    accident detection in first-person (dashboard-mounted camera) videos.
    Our major novelty is to detect anomalies by predicting the future
    locations of traffic participants and then monitoring the
    prediction accuracy and consistency metrics with three different
    strategies. We evaluate our approach using a new dataset of diverse traffic
    accidents, AnAn Accident Detection (A3D), as well as another
    publicly-available dataset. Experimental results show that our approach
    outperforms the state-of-the-art. \textit{Code and the dataset developed in this work are available at:
    \url{https://github.com/MoonBlvd/tad-IROS2019}}
\end{abstract}

\section{Introduction}

Autonomous driving has the potential to transform the world as we know
it, revolutionizing transportation by making it faster,
safer, cheaper, and less labor intensive. A key challenge is building
autonomous systems that can accurately perceive and
safely react to the huge diversity in situations that are encountered
on real-world roadways. Driving situations obey
a long-tailed distribution, such that a very small number of common
situations makes up the vast majority of what a driver encounters, and
a virtually infinite number of rare scenarios --- animals running into
the roadway, cars driving on the wrong side of the street, etc. ---
makes up the rest. While each of these individual scenarios is rare,
they can and do happen. In fact, the chances that \textit{one} of them
will occur on any given day are actually quite high.

Existing work in computer vision has applied deep learning-based visual
classification to detect action starts and their associated
categories~\cite{gao2019startnet} in the video collected by dashboard-mounted
cameras~\cite{chan2016anticipating}. The
long-tailed distribution of driving events means that unusual events
may occur so infrequently that it may be impossible to collect
training data for them, or to even anticipate that
they might occur~\cite{liu2018future}.
In fact, some studies indicate that driverless cars would need to be
tested for \textit{billions} of miles before enough of these rare
situations occur to even accurately measure system
safety~\cite{kalra2016driving}, much less to collect sufficient
training data to make them work well.

\begin{figure}
    \centering
    \includegraphics[width=0.49\textwidth]{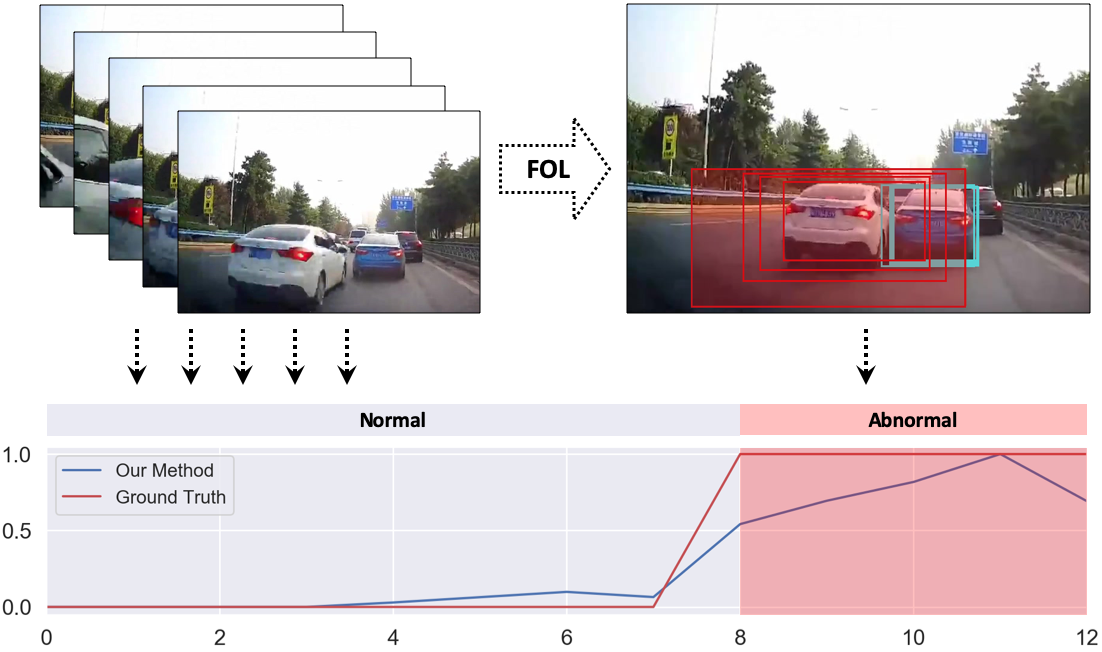}
    \caption{
        Overview of our proposed approach. For each time $t$, we monitor
        the accuracy and consistency of all traffic participants'
        predicted bounding boxes from previous frames and
        calculate the scene's anomaly score.
    }
    \vspace{-5pt}
    \label{fig:teaser}
    \vspace{-10pt}
\end{figure}

An alternative approach is to avoid modeling all possible driving
scenarios, but instead to train models that recognize ``normal,''
safe roadway conditions, and then signal an anomaly when events that
do not fit the model are observed.
Unlike the fully-supervised classification-based work, this unsupervised 
approach would not be able to identify exactly which anomaly has occurred,
but it may still provide sufficient information for the driving system to
recognize an unsafe situation and take evasive action.
This paper proposes a novel approach that learns a deep neural
network model to predict the future locations of objects such as
cars, bikes, pedestrians, etc., in the field of view of a
dashboard-mounted camera on a moving ego-vehicle. 
These models can be easily learned from massive collections of dashboard-mounted
video of normal driving, and no manual labeling is required.
We then compare predicted object locations to the actual locations observed in
the next few video frames. We hypothesize that anomalous roadway
events can be detected by looking for major deviations between the
predicted and actual locations, because unexpected roadway events
(such as cars striking other objects) result in sudden
unexpected changes in an object's speed or position.

Perhaps the closest related work to ours is Liu~\etal~\cite{liu2018future}, who also detect
anomalous events in video. Their technique tries to predict entire future RGB frames and then 
looks for deviations between those and observed RGB frames.
But while their approach can work well for static cameras, 
accurately predicting whole frames is extremely difficult when cameras
are rapidly moving, as in the driving scenario.
We side-step this difficult problem by detecting objects and predicting
their trajectories, as opposed to trying to predict
whole frames. To model the moving camera, 
we explicitly predict the
future odometry of the ego-vehicle; this also allows us to detect significant deviations of the predicted
and real ego-motion, which can be used to 
classify if the ego-vehicle 
is involved in the accident or is just an observer.
We evaluate our technique in extensive experiments on three datasets,
including a new labeled dataset of some 1,500 video traffic accidents
from dashboard cameras that we collected from YouTube. We find that
our method significantly outperforms a number of baselines, including
the published state-of-the-art in anomaly detection.

\section{Related Work}

\xhdr{Trajectory Prediction.}
Extensive research has investigated trajectory prediction, often
posed as a sequence-to-sequence generation problem.
Alahi~\etal~\cite{Alahi_2016_CVPR} introduce a Social-LSTM for
pedestrian trajectories and their interactions. The proposed
social pooling method is further improved by
Gupta~\etal\cite{gupta2018social} to capture global context in
a Generative Adversarial Network (GAN). Social pooling is also applied
to vehicle trajectory prediction in Deo~\etal~\cite{Deo2018} with
multi-modal maneuver conditions. Other
work~\cite{sadeghian2018car,sadeghian2018sophie} captures scene context
information using attention mechanisms to assist trajectory prediction.
Lee~\etal~\cite{lee2017desire} incorporate Recurrent Neural Networks
(RNNs) with conditional variational autoencoders (CVAEs) to generate
multimodal predictions and choose the best by ranking scores. 

While the above methods are designed for third-person views
from static cameras, recent work has considered vision in
first-person (egocentric) videos that capture the natural field of
view of the person or agent (e.g., vehicle) wearing the
camera to study the
camera wearer's actions~\cite{li2015delving,ma2016going},
trajectories~\cite{bertasius2018egocentric},
interactions~\cite{fan2017identifying,xu2018joint}, etc.
Bhattacharyya~\etal~\cite{Bhattacharyya_2018_CVPR} predict
future locations of pedestrians from vehicle-mounted cameras,
modeling observation
uncertainties with a Bayesian LSTM network.
Yagi~\etal~\cite{Yagi_2018_CVPR} incorporate different kinds of cues
into a convolution-deconvolution (Conv1D) network to predict
pedestrians' future locations. Yao~\etal~\cite{yao2018egocentric}
extend this work to autonomous driving scenarios by proposing a
multi-stream RNN Encoder-Decoder (RNN-ED) architecture with both past
vehicle locations and image features as inputs for anticipating
vehicle locations.

\xhdr{Video Anomaly Detection.}  Video anomaly detection has received
considerable attention in computer vision and
robotics~\cite{chandola2009anomaly}. Previous work mainly focuses on
video surveillance scenarios typically using an unsupervised learning
method on the reconstruction of normal training data. For example,
Hasan~\etal~\cite{hasan2016learning} propose a 3D convolutional
Auto-Encoder (Conv-AE) to model non-anomalous frames. To take 
advantage of temporal information,
\cite{medel2016anomaly,chong2017abnormal} use a Convolutional LSTM
Auto-Encoder (ConvLSTM-AE) to capture regular visual and motion
patterns simultaneously. Luo~\etal~\cite{luo2017revisit} propose a
special framework of sRNN, called temporally-coherent sparse coding
(TSC), to preserve the similarities between frames within normal and
abnormal events. Liu~\etal~\cite{liu2018future} detect anomalies by
looking for differences between a predicted future frame and the
actual frame. However, in dynamic autonomous driving scenarios, it is
hard to reconstruct either the current or future RGB frames due to the
ego-car's intense motion. It is even harder to detect abnormal
events. This paper proposes detecting
accidents on roads by using the difference between predicted and actual
trajectories of other vehicles. Our method not only eliminates the
computational cost of reconstructing full RGB frames, but also 
localizes potential anomaly participants.

Prior work has also detected anomalies such as moving violations and
car collisions on roads. Chan~\etal~\cite{chan2016anticipating}
introduce a dataset of crowd-sourced dashcam videos and a
dynamic-spatial-attention RNN model for accident detection.
Herzig~\etal~\cite{herzig2018classifying} propose a Spatio-Temporal
Action Graph (STAG) network to model the latent graph structure of spatial and temporal relations between objects. These
methods are based on supervised learning that requires
arduous human annotations and makes the unrealistic assumption that all
abnormal patterns have been observed in the training data.
This paper considers the challenging but practical problem
of predicting accidents with unsupervised learning.
To evaluate our approach, we introduce a new dataset with traffic
accidents involving objects such as cars and pedestrians.

\section{Unsupervised Traffic Accident Detection\\in First-Person Videos}

Autonomous vehicles must monitor the roadway ahead for signs of
unexpected activity that may require evasive action. A natural way to detect these anomalies is to
look for unexpected or rare movements in the
first-person perspective of a front-facing, dashboard-mounted camera
on a moving ego-vehicle. Prior work~\cite{liu2018future}
proposes monitoring for unexpected scenarios by using past video frames to
predict the current video frame, and then comparing it to the
observed frame and looking for major differences. However, this does
not work well for moving cameras on vehicles, where the perceived
optical motion in the frame is induced by both moving objects and camera
ego-motion. More importantly, anomaly detection systems do \emph{not} need
to accurately predict all information in the frame, since anomalies
are unlikely to involve peripheral objects such as houses or
billboards by the roadside. This paper thus assumes that an anomaly may exist if an
object's real-world observed trajectory deviates from the predicted
trajectory. For example, when a vehicle should move through
an intersection but instead suddenly stops, a collision may have occurred.

Following Liu~\etal~\cite{liu2018future}, our model is trained with a
large-scale dataset of normal, non-anomalous driving videos. This
allows the model to learn normal patterns of object and ego motions,
 then recognize deviations without the need to
explicitly train the model with examples of every possible anomaly. This
video data is easy to obtain and does not require hand labeling.
Considering the influence of ego-motion on perceived object location, we incorporate a future ego-motion
prediction module~\cite{yao2018egocentric} as an additional input.
At test time, we use the model to predict the current locations of
objects based on the last few frames of data and determine if an
abnormal event has happened based on three different anomaly detection
strategies, as described in Section~\ref{sec:metrics}.

\begin{figure}
    \vspace{5pt}
    \centering
    \includegraphics[width=0.48\textwidth]{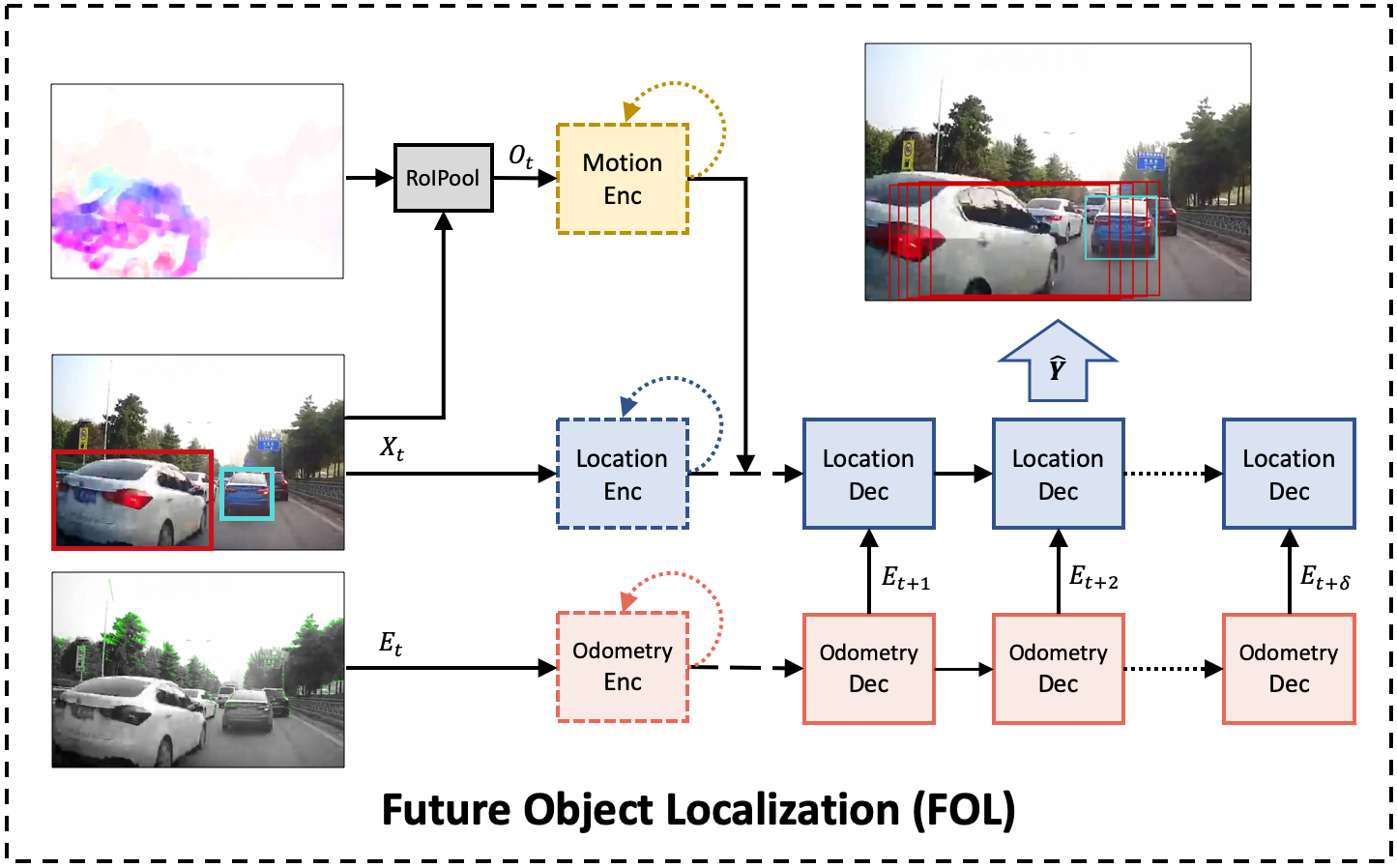}
    \caption{
        Overview of the future object localization model.
    }
    \label{fig:fol}
    \vspace{-15pt}
\end{figure}

\subsection{Future Object Localization (FOL)}

\subsubsection{Bounding Box Prediction}
Following ~\cite{yao2018egocentric}, we denote an
observed object's bounding box  $X_t=[c^x_t,c^y_t,w_t,h_t]$ at
time $t$, where ($c^x_t$, $c^y_t$) is the location of the center of
the box and $w_t$ and $h_t$ are its width and height in pixels,
respectively. We denote the object's future bounding box trajectory
for the $\delta$ frames after time $t$ to be
$\mathbf{Y_t}=\{Y_{t+1},Y_{t+2},\cdots,Y_{t+\delta}\}$, where each
$Y_t$ is a bounding box parameterized by center, width, and height.
Given the image evidence $O_t$ observed at time $t$, a visible
object's location $X_t$, and its corresponding historical information
$H_{t-1}$, our future object localization model predicts
$\mathbf{Y_t}$. This model is
inspired by the
multi-stream RNN encoder-decoder framework of Yao
\etal~\cite{yao2018egocentric}, but with completely different network
structure~\cite{xu2018temporal}.
For each frame,~\cite{yao2018egocentric} receives and
re-processes the previous 10 frames
before making a decision, whereas
our model only needs to process the current information, making it
much faster at inference time. Our model is shown in 
Figure~\ref{fig:fol}.
Two encoders (Enc) based on gated recurrent units (GRUs) receive
an object's current bounding box and pixel-level spatiotemporal
features as inputs, respectively, and update the object's hidden
states. In particular, the spatiotemporal 
features are extracted by a
region-of-interest pooling (RoIPool) operation using bilinear
interpolation from precomputed optical flow fields. 
The updated hidden states are used by a
location decoder (Dec) to recurrently predict the bounding boxes
of the immediate future.

\subsubsection{Ego-Motion Cue}
Ego-motion information of the moving camera has been shown
to be necessary for accurate future object localization~\cite{yao2018egocentric,Bhattacharyya_2018_CVPR}.
Let $E_t$ be the ego-vehicle's pose at time $t$;
$E_t=\{\phi_t,x_t,z_t\}$ where $\phi_t$ is the yaw angle and $x_t$
and $z_t$ are the positions along the ground plane with respect to the
vehicle's starting position in the first video frame.
We predict the ego-vehicle's odometry by using another RNN 
encoder-decoder module to encode  ego-position
change vector $E_t-E_{t-1}$ and decode future ego-position changes
$\mathbf{E}=\{\hat{E}_{t+1}-E_{t},\hat{E}_{t+2}-E_{t},...,\hat{E}_{t+\delta}-E_{t}\}$.
We use the change in ego-position 
to eliminate accumulated odometry errors. 
The output $\mathbf{E}$ is then combined with the hidden state of the
future object localization decoder to form the input into the next time step.

\subsubsection{Missed Objects}
We build a list of trackers $Trks$ per~\cite{wojke2017simple} to 
record the current bounding box
$Trks[i].X_t$, the predicted future boxes $Trks[i].\mathbf{\hat{Y}}_t$, 
and the tracker
age $Trks[i].age$ of each object. 
We denote all maintained track IDs as $D$ (both observed 
and missed), all currently observed track IDs as $C$, and the 
missed object IDs as $D-C$.
At each time step, we
update the observed trackers and initialize a new tracker when a new object
is detected.  For objects that are temporarily missed (i.e., occluded), 
we use their previously predicted bounding boxes as
their estimated current location and run future object localization 
with RoIPool features
from those predicted boxes per Algorithm~\ref{alg:fol_track}. 
This missed object mechanism is essential in our prediction-based
anomaly detection method to eliminate the impact of failed object
detection or tracking in any given frame. For example, if an object with a normal motion
pattern is missed for several frames, the FOL is still expected to
give reasonable predictions except for some accumulated deviations. On the
other hand, if an anomalous object is missed during tracking~\cite{wojke2017simple}, 
FOL-Track will make a prediction using its previously
predicted bounding box whose region can be totally displaced and can
result in inaccurate predictions. In this case, some false alarms and false
negatives can be eliminated by using the metrics presented in
Section~\ref{sec:metrics_three}.

\vspace{-4pt}
\begin{algorithm}
    \SetKwInOut{Input}{Input}
    \SetKwInOut{Output}{Output}
    \Input{Observed bounding boxes $\{X_t^{(i)}\}$ where $i \in C$,
    observed image evidence $O_t$, trackers of all objects $Trks$ with track IDs $D$}
    \Output{Updated trackers $Trks$}
    $A$ is the maximum age of a tracker \\
	\For(\tcp*[f]{update observed trackers}){$i\in C$}{
	    \uIf{$i \notin D$}{
            initialize $Trks[i]$ \\
	    }
	    \Else{
            $Trks[i].X_t=X_t^{(i)}$ \\
            $Trks[i].\mathbf{\hat{Y}}_t=FOL(X_t^{(i)}, O_t)$
        }
	}
	\For(\tcp*[f]{update missed trackers}){$j\in D - C$}{
		\uIf{$Trks[j].age > A$}{
            remove $Trks[j]$ from $Trks$ \\
		}
		\Else{
            $Trks[j].X_t=Trks[j].\hat{Y}_{t-1}$ \\
            $Trks[j].\mathbf{\hat{Y}}_t=FOL(Trks[j].X_t, O_t)$ 
		}
	}		   
	\caption{FOL-Track Algorithm}
	\label{alg:fol_track}
\end{algorithm}
\vspace{-8pt}

\begin{figure*}
    \vspace{5pt}
    \centering
    \includegraphics[width=0.8\textwidth]{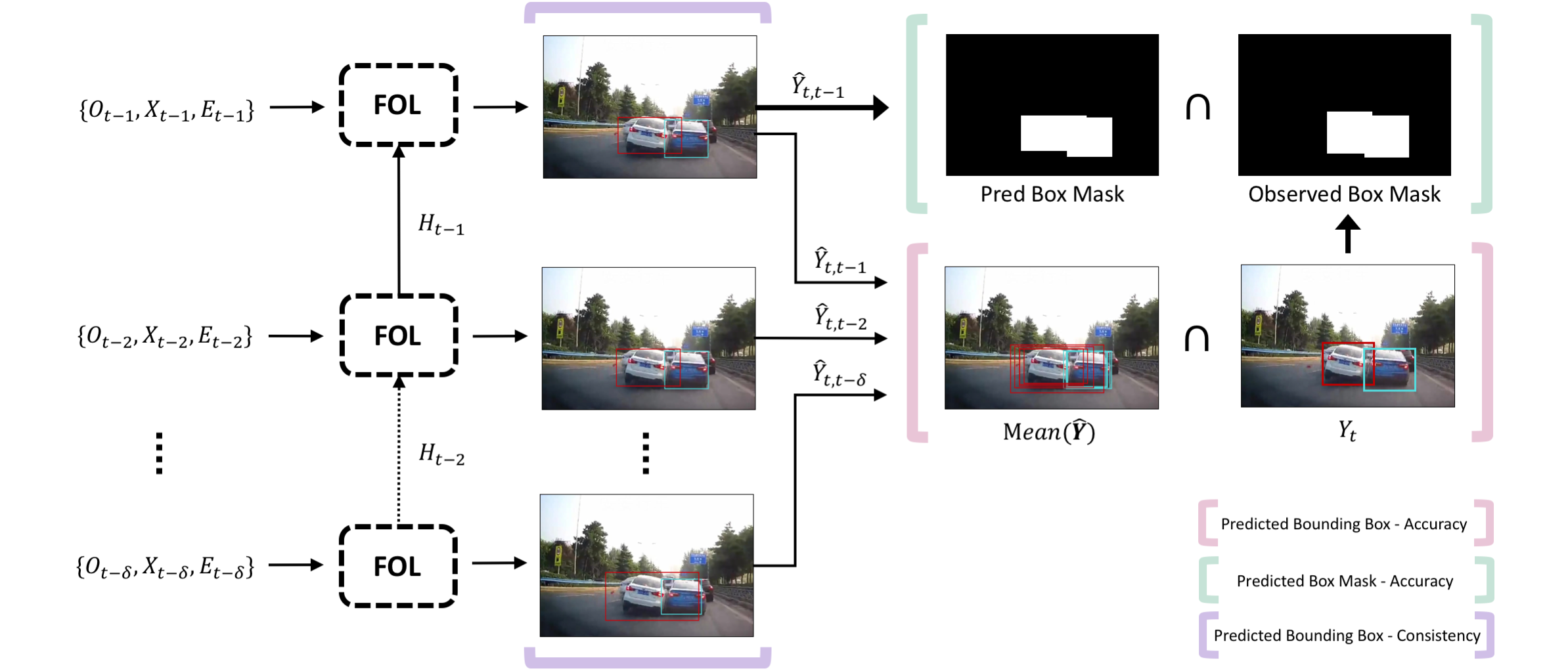}
    \caption{
        Overview of our unsupervised traffic accident detection methods. The three brackets correspond to: (1) Predicted bounding box accuracy method (pink); (2) Predicted box mask accuracy method (green); (3) Predicted bounding box consistency method (purple). All methods use multiple previous FOL outputs to compute anomaly scores.
    }
    \label{fig:metrics}
    \vspace{-12pt}
\end{figure*}

\subsection{Traffic Accident Detection}
\label{sec:metrics}
In this section, we propose three different strategies for traffic
accident detection by monitoring the prediction accuracy and consistency
of objects' future locations. The key idea is that object trajectories
and locations in non-anomalous events can be precisely predicted,
while deviations from predicted behaviors suggest an anomaly.

\subsubsection{Predicted Bounding Boxes - Accuracy}
\label{sec:metrics_one}
One simple method for recognizing abnormal
events is to directly measure the similarity between predicted
object bounding boxes and their corresponding observations.
The FOL model predicts bounding boxes
of the next $\delta$ future frames, i.e., at each time $t$ each object has $\delta$ bounding boxes
predicted from time $t-\delta$ to $t-1$, respectively.
We first average the positions of the $\delta$ bounding boxes,
then compute intersection over union (IoU) between the averaged bounding box and the observed box location,
where higher IoU means greater agreement between the two boxes.
We average computed IoU values over all observed
objects and then compute an aggregate anomaly score $L_{bbox}\in[0,1]$,
\begin{equation}\label{eq:l_bbox}
    L_{bbox} = 1 - \frac{1}{N}\sum_{i=1}^{N}
    \text{\textbf{IoU}}\bigg(\Big(\frac{1}{\delta}\sum_{j=1}^{\delta}\hat{Y}_{t, t-j}^{i}\Big), Y_{t_0}^{i}\bigg),
\end{equation}
where $N$ is the total number of observed objects, and $\hat{Y}_{t,t-j}^{i}$
is the predicted bounding box from time $t-j$ of object $i$ at time $t$.
This method relies upon accurate object tracking to match the predicted
and observed bounding boxes.

\subsubsection{Predicted Box Mask - Accuracy}
\label{sec:metrics_two}
Although tracking algorithms such as Deep-SORT~\cite{wojke2017simple}
offer reasonable accuracy, it is still possible
to lose or mis-track objects. We found that inaccurate tracking
particularly happens in severe traffic accidents because of the twist and distortion of
object appearances. Moreover, severe ego-motion also results in
inaccurate tracking due to sudden changes in object 
locations. This increases the number of false negatives of the metric
proposed above, which simply ignores objects that are not
successfully tracked in a given frame.
To solve this problem, 
we first convert all areas within the predicted bounding boxes to
binary masks, with areas inside the boxes having value $1$ and backgrounds having $0$,
and do the same with the observed boxes.
We then calculate an anomaly score as the IoU between these two binary masks,
\begin{align}
    & I^{(u,v)}  = \begin{cases}
                    1 \text{, if pixel $(u,v)$ within box $X^{i},\, \forall i,$}\\
                    0 \text{, otherwise,}
                \end{cases} \label{eq:mask} \\
    & L_{mask} = 1 - \text{\textbf{IoU}}\big(\hat{I}_{t,t-1}, I_t\big), \label{eq:l_mask}
\end{align}
where $I^{(u,v)}$ is pixel $(u,v)$ on mask $I$, $X^{i}$ is the
$i$-th bounding box, $\hat{I}_{t,t-1}$ is the predicted mask
from time $t-1$, and $I_t$ is the observed mask at $t$.
In other words, while the metric in the last section compares bounding
boxes on an object-by-object basis, this metric simply compares the bounding
boxes of all objects simultaneously.
The main idea is that
accurate prediction results will still have a relatively large IoU compared
to the ground truth observation.

\begin{table*}[h]
    \vspace{5pt}
    \centering
    \renewcommand{\arraystretch}{1.3}
    \caption{Comparison of publicly available datasets for video anomaly detection.
    $^*$Surveillance videos. $^{**}$Egocentric videos (training frames are all normal videos, while some test frame videos contain anomalies.) }
    \label{tab:dataset}
    \begin{tabular}{l|c|c|c|c|c}
        \toprule
        Dataset & \# videos & \# training frames & \# testing frames & \# anomaly events & typical participants \\
        \midrule
        UCSD Ped1/Ped2$^*$~\cite{li2014anomaly} & 98 & 9,350 & 9,210 & 77 & bike, pedestrian, cart, skateboard \\
        CUHK Avenue$^*$~\cite{lu2013abnormal} & 37 & 15,328 & 15,324 & 47 & bike, pedestrian  \\
        UCF-Crime$^*$~\cite{Sultani_2018_CVPR} & 1,900 &1,610 videos & 290 videos & 1,900 & car, pedestrian, animal \\
        ShanghaiTech$^*$~\cite{luo2017revisit} & 437 & 274,515 & 42,883 & 130 & bike, pedestrian \\
        Street Accidents (SA)$^{**}$~\cite{chan2016anticipating} & 994 & 82,900 & 16,500 & 165 & car, truck, bike \\
        \textbf{A3D}$^{**}$ & \textbf{1,500} & \textbf{79,991} (HEV-I) & \textbf{128,175} & \textbf{1,500} & \textbf{car, truck, bike, pedestrian, animal}  \\
        \bottomrule
    \end{tabular}
    \vspace{-15pt}
\end{table*}

\subsubsection{Predicted Bounding Boxes - Consistency}
\label{sec:metrics_three}
The above methods rely on accurate detection of objects in consecutive
frames to compute anomaly scores. However, the detection of anomaly
participants is not always accurate due to changes in appearance
and mutual occlusions. We hypothesize that visual and motion features
about an anomaly do not only appear once it happens, but usually are
accompanied by a salient pre-event. We thus propose another strategy
to detect anomalies by computing consistency of future object localization
outputs from several previous frames while eliminating the effect of
inaccurate detection and tracking.

As discussed in Section~\ref{sec:metrics_one}, our model has $\delta$
predicted bounding boxes for each object in video frame $t$.
We compute the standard deviation (STD) between all $\delta$
predicted bounding boxes to measure their similarity,
\begin{equation}\label{eq:l_pred}
    L_{pred} = \frac{1}{N}\sum_{i=1}^{N} \max_{\{c^x,c^y,w,h\}} \text{\textbf{STD}}(\hat{Y}_{t,t-j}).
\end{equation}
We compute the maximum STD over the four components of
the bounding boxes since different anomalies may be indicated by
different effects on the bounding box, e.g., suddenly stopped
cross traffic
may only have large STD along the horizontal axis.
A low STD suggests the object is following normal
movement patterns and thus the predictions
are stable, while a high standard deviation suggests abnormal motion.
For all three methods, we follow \cite{liu2018future} to
normalize computed anomaly scores for evaluation.

\section{Experiments}

%\subsection{Dataset}
To evaluate our method on realistic traffic scenarios, we introduce a
new dataset, \textbf{AnAn Accident Detection (A3D)}, of on-road abnormal 
event videos compiled as 1500 video clips from a YouTube 
channel~\cite{ananxingche} of dashboard cameras from different cars in East Asia.
Each video contains an abnormal traffic event at different temporal
locations. We labeled each video with anomaly start and end times
under the consensus of three human annotators. The
annotators were instructed to label the anomalies based on common sense, with the start
time defined to be the point when the accident is inevitable and the end time the point
when all participants recover a normal \textit{moving} condition or fully stop. 

We compare our A3D dataset with existing video anomaly detection
datasets in Table~\ref{tab:dataset}. A3D includes a total
of 128,175 frames (ranging from 23 to 208 frames)
at 10 frames per second and is clustered into 18 types
of traffic accidents each labeled with a brief description.
A3D includes driving scenarios with different weather conditions
(e.g., sunny, rainy, snowy, etc.), places
(e.g., urban, countryside, etc.), and participant types
(e.g., cars, motorcycles, pedestrians, animals, etc.).
In addition to start and end times, each traffic anomaly is 
labeled with a binary value indicating whether the ego-vehicle is involved,
to provide a better
understanding of the event. Note that this could especially benefit the
first-person vision community. For example, rear-end collisions are
the most difficult to detect from traditional
anomaly detection methods. About $60\%$ of accidents in the dataset involve the
ego-vehicle, and others are observed by moving cars from
a third-person perspective.

Since A3D does not contain nominal videos, we use the publicly
available Honda Egocentric View Intersection
(HEV-I)~\cite{yao2018egocentric} dataset to train our model. HEV-I
was designed for future object localization and consists of 230
on-road videos at intersections in the San Francisco Bay Area. Each video 
is 10-60 seconds in length. Since HEV-I and A3D
were collected in different places with different kinds of cameras,
there is no overlap between the training and testing datasets.
Following prior work~\cite{yao2018egocentric}, we produce object
bounding boxes using Mask-RCNN~\cite{he2017mask} pre-trained on the
COCO dataset and find tracking IDs using
Deep-SORT~\cite{wojke2017simple}.

\subsection{Implementation Details}
We implemented our model in PyTorch~\cite{pytorch} and performed experiments 
on a system with an Nvidia Titan Xp Pascal
GPU. We use ORB-SLAM 2.0~\cite{mur2017orb} for ego
odometry calculation and compute optical flow using FlowNet
2.0~\cite{ilg2017flownet}. In our training data (HEV-I), 
we used the provided camera intrinsic matrix. We used the same matrix 
for A3D and SA
since these videos are collected from different dash cameras and the 
parameters are unavailable. We also set the feature count to $12000$ to 
have a better performance.
We use a 5$\times$5 RoIPool operator to
produce the final flattened feature vector $O_t \in
\mathbb{R}^{50}$. The gated recurrent unit
(GRU)~\cite{chung2015gated} is our basic RNN cell. GRU hidden state sizes 
for future object localization  and the ego-motion prediction model were
set to 512 and 128, respectively. To learn network parameters, we
use the RMSprop~\cite{hinton2012neural} optimizer with default
parameters, learning rate $10^{-4}$, and no weight decay. Our models
were optimized in an end-to-end manner, and the training process was
terminated after 100 epochs using a batch size of 32. The best model
was selected according to its performance in future object localization.

\subsection{Evaluation Metrics}

For accident detection evaluation, we follow the literature of video anomaly 
detection~\cite{li2014anomaly}
and compute frame-level Receiver Operation Characteristic (ROC) curves and Area
Under the Curve (AUC). A higher AUC value indicates better
performance.

\subsection{Video Anomaly Detection Baselines}
\xhdr{\textit{K}-Nearest Neighbor Distance.}
We segment each video into a bag of short video chunks of 16 frames. Each 
chunk is labeled as either normal or anomalous based on the
annotation of the 8-th frame. We then feed each chunk into an I3D~\cite{carreira2017quo}
network pre-trained on Kinetics dataset, and extract the outputs
of the last fully connected layer as its feature representations.
All videos in the HEV-I dataset are used as normal data. The normalized distance
of each test video chunk to the centroid of its $K$ nearest normal
($K$-NN) video chunks is computed as the anomaly score. We show
results of $K=1$ and $K=5$ in this paper.

\xhdr{Conv-AE~\cite{hasan2016learning}.} We reimplement the Conv-AE
model for unsupervised video anomaly detection by
following~\cite{hasan2016learning}.
The input images are encoded by 3
convolutional layers and 2 pooling layers, and then decoded by 3
deconvolutional layers and 2 upsampling layers for
reconstruction.
Anomaly score computation is from~\cite{hasan2016learning}.
The model is trained on a mixture of the SA (Table~\ref{tab:dataset})
and the HEV-I dataset for 20 epochs and the best
model is selected.

\xhdr{State-of-the-art~\cite{liu2018future}.}
The future frame prediction network with Generative Adversarial Network (GAN)
achieved the state-of-the-art results for video anomaly detection.
This work detects abnormal events by
leveraging the difference between a predicted future frame and its
ground truth. To fairly compare with our method, we used
the publicly available code by the authors of~\cite{liu2018future} and finetuned on the same
dataset as Conv-AE. Training is terminated after 100,000 iterations
and the best model is selected.

\subsection{FOL Results}

\begin{table}[t]
    \vspace{5pt}
    \centering
    \renewcommand{\arraystretch}{1.3}
    \caption{Experimental results of FOL (errors are in pixels).}
    \label{tab:fol_result}
    \begin{tabular}{lcccc}
        \toprule
        Dataset  & Prediction Horizon & FDE  & ADE  & FIOU \\
        \midrule 
        HEV-I (test)~\cite{yao2018egocentric} & 0.5 sec & 11.0 & 6.7 & 0.85 \\
        SA (test)~\cite{chan2016anticipating} & 0.5 sec & 21.3 & 13.5 & 0.64 \\
        A3D & 0.5 sec & 25.6 & 16.4 & 0.63 \\
        \bottomrule
    \end{tabular}
    \vspace{-5pt}
\end{table}

We first show the performance of the pretrained FOL model on HEV-I's
validation set and on the other two accident datasets (SA and A3D).
Similar to~\cite{yao2018egocentric}, 
the final displacement error (FDE), 
average displacement error (ADE), and final IOU (FIOU) are presented 
in Table~\ref{tab:fol_result}. The FDEs and ADEs on A3D 
and SA are higher and the FIOUs are lower than HEV-I because these 
videos were collected 
from different dash cameras in different scenarios, while all HEV-I 
videos were collected using the same cameras. 
The accidents in these videos result in 
lower FOL prediction accuracy, which is consistent with the 
assumption of our proposed approach.
The overall FOL performance on A3D is slightly 
worse compare to SA since A3D is a larger dataset with more diverse 
accident types.

\subsection{Accident Detection Results on A3D Dataset}

\begin{table}[t]
    \vspace{5pt}
    \centering
    \renewcommand{\arraystretch}{1.3}
    \caption{Experimental results on A3D and SA datasets in terms of AUC.}
    \label{tab:results}
    \begin{tabular}{lccc}
        \toprule
        Methods & A3D & A3D (w/o Ego) & SA~\cite{chan2016anticipating} \\
        \midrule
        % Euclidean Distance  & 48.6 & 50.8 & 47.6  \\
        $K$-NN (K = 1)  & 48.0 & 51.3 & 48.2 \\
        $K$-NN (K = 5)  & 47.8 & 51.2 & 48.1 \\
        Conv-AE\cite{hasan2016learning} & 49.5 & 49.9 & 50.4 \\
        State-of-the-art~\cite{liu2018future}  & 46.1  & 50.7 & 50.4 \\
        \midrule
        FOL-AvgIoU   & 49.7 & 57.0 & 53.4  \\
        FOL-MinIoU   & 48.4 & 56.0 & 52.6 \\
        FOL-Mask  & 54.1 & 54.9 & 54.8 \\
        FOL-AvgSTD (pred only)   & 59.3 & \textbf{60.2} & \textbf{55.8} \\
        FOL-MaxSTD (pred only)   & \textbf{60.1} & 59.8 & 55.6 \\
        \bottomrule
    \end{tabular}
    \vspace{-10pt}
\end{table}

\vspace{-5pt}
\xhdr{Quantitative Results.}
We evaluated baselines, a
state-of-the-art method, and our proposed method on the A3D dataset.
As shown in the first column of Table~\ref{tab:results}, our
method outperforms the $K-$NN baseline as well as Conv-AE and the state-of-the-art. As a comparative study, we evaluate
performance of our future object localization (FOL) methods with the three metrics presented in
Section~\ref{sec:metrics}. \textit{FOL-AvgIoU} uses the metrics in
Eq.~\eqref{eq:l_bbox}, while \textit{FOL-MinIoU} is a variation where we
evaluate minimum IoU over all observed objects instead of computing
the average, resulting in not only anomaly detection but also
anomalous object localization. However, \textit{FOL-MinIoU} can
perform worse since it is not robust to outliers such as failed
prediction of a normal object, which is more frequent in videos with a
large number of objects. \textit{FOL-Mask} uses the metrics in
Eq.~\eqref{eq:l_mask} and significantly outperforms the above two methods. This
method does not rely on accurate tracking, so it handles cases
including mis-tracked objects. However, it may mis-label a frame as an
anomaly if object detection loses some normal objects. Our best
methods use the prediction-only metric defined in
Eq.~\eqref{eq:l_pred} which has two variations \textit{FOL-AvgSTD}
and \textit{FOL-MaxSTD}. Similar to the IoU based methods, 
\textit{FOL-MaxSTD} finds the most anomalous object in the frame.
By using only prediction, our method is insensitive to
unreliable object detection and tracking when an anomaly happens,
including the false negatives (in IoU based methods) and the false
positives (in Mask based methods) caused by losing 
objects. However, this method can fail in cases where predicting future locations
of an object is difficult, e.g., an object with low resolution,
intense ego-motion, or multiple object occlusions due to heavy
traffic.

We also evaluated the methods by removing videos where ego cars 
(A3D w/o Ego in Table~\ref{tab:results})
are involved in anomalies to show how ego motion influences
anomaly detection performance. As shown in the second and the third
columns of Table~\ref{tab:results}, \textit{FOL-AvgIoU} and
\textit{FOL-MinIoU} perform better on videos where ego camera is steady
while the
other methods are relatively robust to ego-motion. This further
shows that it is necessary to reduce dependency on accurate object
detection and tracking when anomalies occur.

\begin{figure*}
    \vspace{5pt}
    \center
    \begin{subfigure}[htb]{1.0\textwidth}
        \center
        \includegraphics[width=0.96\linewidth]{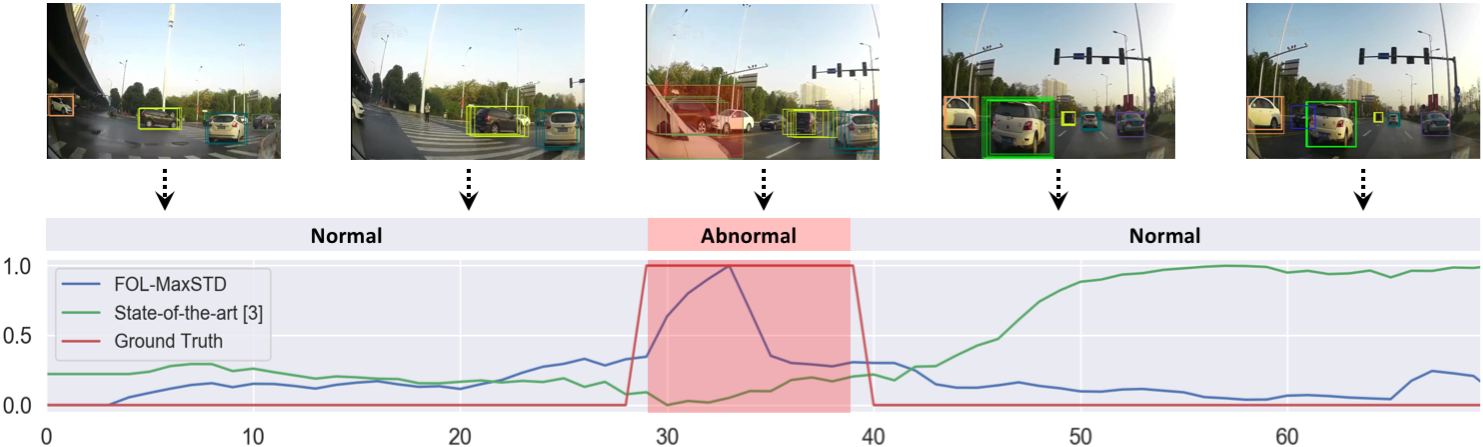}
        \vspace{1pt}
        \label{fig:result_1}
    \end{subfigure}
    \begin{subfigure}[htb]{1.0\textwidth}
        \center
        \includegraphics[width=0.96\linewidth]{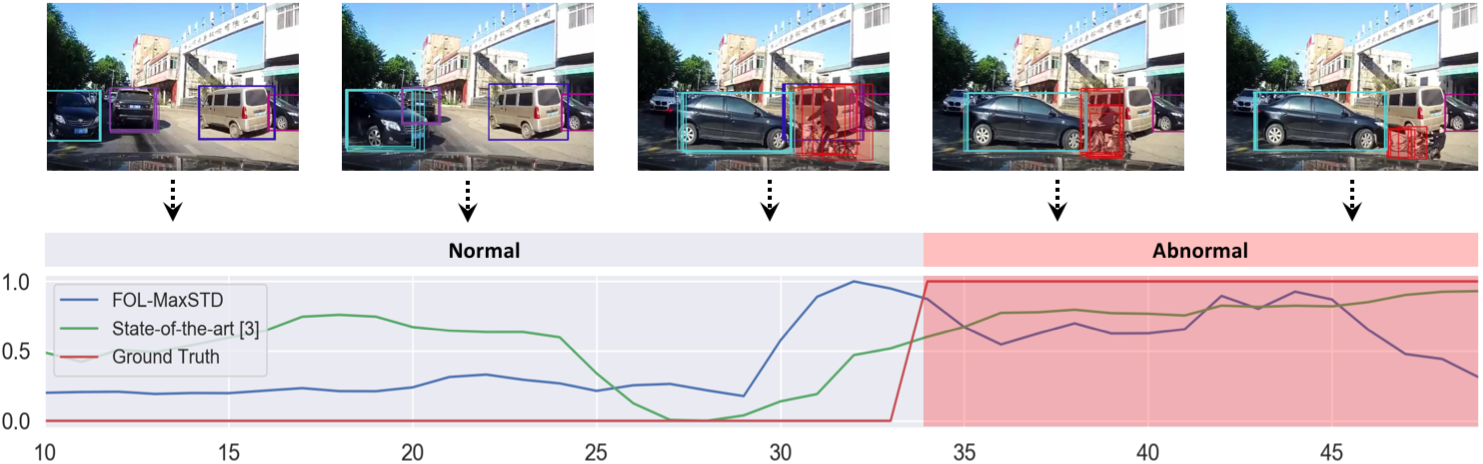}
        \vspace{1pt}
        \label{fig:result_2}
    \end{subfigure}
    \caption{
        Two examples of our best method and a state-of-the-art method on the A3D dataset.
        }
    \label{fig:result_1_2}
\end{figure*}

\begin{figure*}[t]
    \center
    \includegraphics[width=0.96\linewidth]{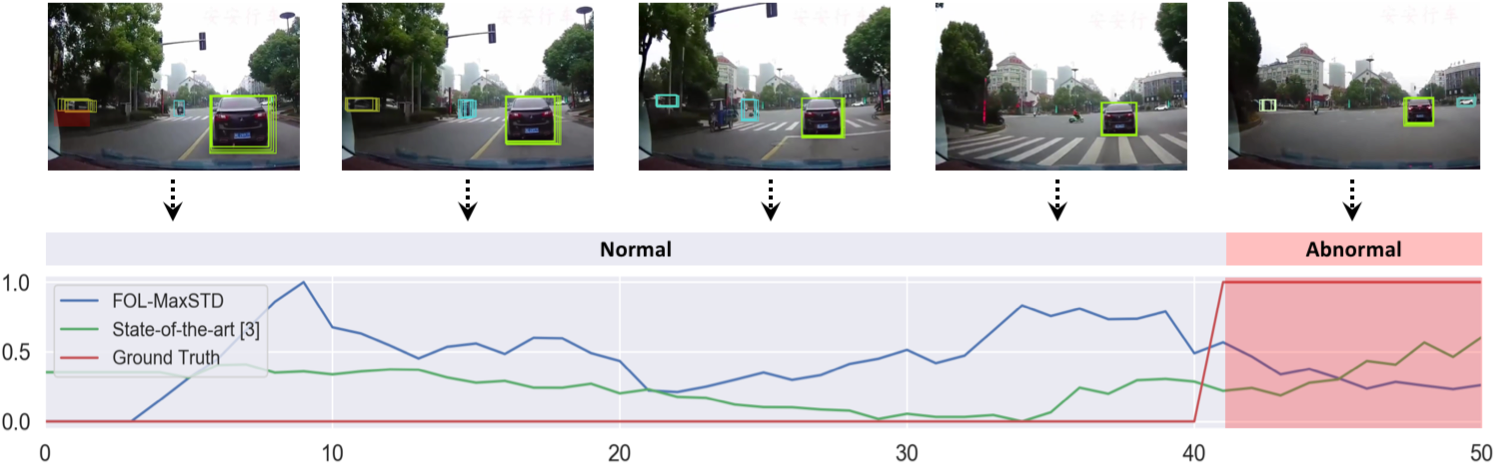}
    \vspace{-2pt}
    \caption{
         A failure case of our method on the A3D dataset with false alarms and false negatives.
    }
    \vspace{-10pt}
    \label{fig:result_4}
\end{figure*}

\xhdr{Qualitative Results.}
Fig.~\ref{fig:result_1_2} shows
two sample results of our best method and the published state-of-the-art on the A3D
dataset. For example, in the upper one, predictions of all observed
traffic participants are accurate and consistent at the beginning.
The ego car is hit at around the $30$-th frame by the white car on its
left, causing inaccurate and unstable predictions and generating
high anomaly scores. After the crash, the ego car stops and the
predictions recover, as presented in the last two images. Fig.~\ref{fig:result_4} 
shows a failure case where our method makes false alarms at the beginning
due to inconsistent prediction of the very left car occluded by trees.
This is because our model takes all objects into consideration equally rather than focusing on important objects. False negatives show that our method is not able to detect an accident
if participants are totally occluded (e.g. the bike) or the motion pattern is accidentally normal from a particular viewpoint (e.g. the middle car).

\begin{figure*}[htb]
    \center
    \includegraphics[width=0.96\linewidth]{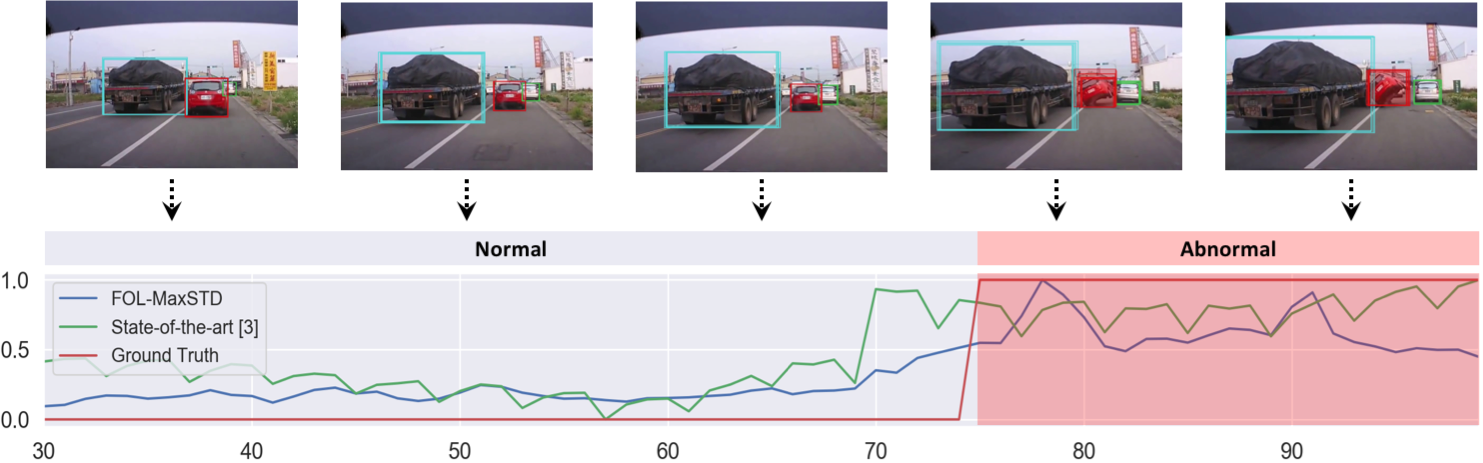}
    \vspace{-2pt}
    \caption{
        An example of our best method and a state-of-the-art method on the SA dataset\cite{chan2016anticipating}.
    }
    \vspace{-10pt}
    \label{fig:result_3}
\end{figure*}

\subsection{Results on the SA Dataset}
We also compared the performance of our model and baselines on the
Street Accident (SA)~\cite{chan2016anticipating} dataset of on-road
accidents in Taiwan. This dataset was collected from
dashboard cameras with 720p resolution from the driver's
point-of-view. Note that we use SA only for testing,
and still train on the HEV-I dataset.
We follow prior work~\cite{chan2016anticipating} and report evaluation results with
165 test videos containing different anomalies.
The right-most column in
Table~\ref{tab:results} shows the results of different methods on SA.
In general, our best method outperforms all baselines and the
published state-of-the-art. The SA testing dataset is much smaller than
A3D,  and we have informally observed that it is biased towards anomalies involving bikes.
It also contains  videos collected from
cyclist head cameras which have irregular camera angles and large
vibrations. Fig.~\ref{fig:result_3} shows an example of anomaly
detection in the SA dataset. 

\vspace{2pt}
\section{Conclusion}
\vspace{2pt}
This paper proposed an unsupervised deep learning framework for
traffic accident detection from egocentric videos. A key challenge is rapid motion of the ego-car,
making visual reconstruction of either current or future RGB frames
from regular training data difficult.
We predicted traffic participant trajectories as well as
their future locations, and utilized anticipation
accuracy and consistency as signals
that an anomaly may have occurred. We introduced a new dataset consisting of a variety of
real-world accidents on roads and also evaluated our method on an existing traffic accident detection dataset.  Experiments showed that our model
significantly outperforms published baselines.

\vspace{2pt}
\section{Acknowledgments}
\vspace{2pt}
This research has been supported by the National Science Foundation
under awards CNS 1544844 and CAREER IIS-1253549,
and by the IU Office of the Vice Provost for Research,
the College of Arts and Sciences, and the School of Informatics,
Computing, and Engineering through the Emerging Areas of Research
Project ``Learning: Brains, Machines, and Children."
The views and conclusions contained in this
paper are those of the authors and should not be interpreted
as representing the official policies, either expressly or implied,
of the U.S. Government, or any sponsor.

\bibliographystyle{IEEEtran}
\bibliography{Reference}

\begin{thebibliography}{10}
\providecommand{\url}[1]{#1}
\csname url@samestyle\endcsname
\providecommand{\newblock}{\relax}
\providecommand{\bibinfo}[2]{#2}
\providecommand{\BIBentrySTDinterwordspacing}{\spaceskip=0pt\relax}
\providecommand{\BIBentryALTinterwordstretchfactor}{4}
\providecommand{\BIBentryALTinterwordspacing}{\spaceskip=\fontdimen2\font plus
\BIBentryALTinterwordstretchfactor\fontdimen3\font minus
  \fontdimen4\font\relax}
\providecommand{\BIBforeignlanguage}[2]{{%
\expandafter\ifx\csname l@#1\endcsname\relax
\typeout{** WARNING: IEEEtran.bst: No hyphenation pattern has been}%
\typeout{** loaded for the language `#1'. Using the pattern for}%
\typeout{** the default language instead.}%
\else
\language=\csname l@#1\endcsname
\fi
#2}}
\providecommand{\BIBdecl}{\relax}
\BIBdecl

\bibitem{gao2019startnet}
M.~Gao, M.~Xu, L.~S. Davis, R.~Socher, and C.~Xiong, ``Startnet: Online
  detection of action start in untrimmed videos,'' \emph{ICCV}, 2019.

\bibitem{chan2016anticipating}
F.-H. Chan, Y.-T. Chen, Y.~Xiang, and M.~Sun, ``Anticipating accidents in
  dashcam videos,'' in \emph{ACCV}, 2016.

\bibitem{liu2018future}
W.~Liu, W.~Luo, D.~Lian, and S.~Gao, ``Future frame prediction for anomaly
  detection--a new baseline,'' in \emph{CVPR}, 2018.

\bibitem{kalra2016driving}
N.~Kalra and S.~M. Paddock, ``Driving to safety: How many miles of driving
  would it take to demonstrate autonomous vehicle reliability?''
  \emph{Transportation Research Part A: Policy and Practice}, 2016.

\bibitem{Alahi_2016_CVPR}
A.~Alahi, K.~Goel, V.~Ramanathan, A.~Robicquet, L.~Fei-Fei, and S.~Savarese,
  ``Social {LSTM}: Human trajectory prediction in crowded spaces,'' in
  \emph{CVPR}, 2016.

\bibitem{gupta2018social}
A.~Gupta, J.~Johnson, L.~Fei-Fei, S.~Savarese, and A.~Alahi, ``Social {GAN}:
  Socially acceptable trajectories with generative adversarial networks,'' in
  \emph{CVPR}, 2018.

\bibitem{Deo2018}
N.~Deo, A.~Rangesh, and M.~M. Trivedi, ``How would surround vehicles move? a
  unified framework for maneuver classification and motion prediction,''
  \emph{T-IV}, 2018.

\bibitem{sadeghian2018car}
A.~Sadeghian, F.~Legros, M.~Voisin, R.~Vesel, A.~Alahi, and S.~Savarese,
  ``{Car-Net}: Clairvoyant attentive recurrent network,'' in \emph{ECCV}, 2018.

\bibitem{sadeghian2018sophie}
A.~Sadeghian, V.~Kosaraju, A.~Sadeghian, N.~Hirose, and S.~Savarese, ``Sophie:
  An attentive gan for predicting paths compliant to social and physical
  constraints,'' \emph{arXiv:1806.01482}, 2018.

\bibitem{lee2017desire}
N.~Lee, W.~Choi, P.~Vernaza, C.~B. Choy, P.~H. Torr, and M.~Chandraker,
  ``Desire: Distant future prediction in dynamic scenes with interacting
  agents,'' in \emph{CVPR}, 2017.

\bibitem{li2015delving}
Y.~Li, Z.~Ye, and J.~M. Rehg, ``Delving into egocentric actions,'' in
  \emph{CVPR}, 2015.

\bibitem{ma2016going}
M.~Ma, H.~Fan, and K.~M. Kitani, ``Going deeper into first-person activity
  recognition,'' in \emph{CVPR}, 2016.

\bibitem{bertasius2018egocentric}
G.~Bertasius, A.~Chan, and J.~Shi, ``Egocentric basketball motion planning from
  a single first-person image,'' in \emph{CVPR}, 2018.

\bibitem{fan2017identifying}
C.~Fan, J.~Lee, M.~Xu, K.~K. Singh, Y.~J. Lee, D.~J. Crandall, and M.~S. Ryoo,
  ``Identifying first-person camera wearers in third-person videos,''
  \emph{CVPR}, 2017.

\bibitem{xu2018joint}
M.~Xu, C.~Fan, Y.~Wang, M.~S. Ryoo, and D.~J. Crandall, ``Joint person
  segmentation and identification in synchronized first-and third-person
  videos,'' \emph{ECCV}, 2018.

\bibitem{Bhattacharyya_2018_CVPR}
A.~Bhattacharyya, M.~Fritz, and B.~Schiele, ``Long-term on-board prediction of
  people in traffic scenes under uncertainty,'' in \emph{CVPR}, 2018.

\bibitem{Yagi_2018_CVPR}
T.~Yagi, K.~Mangalam, R.~Yonetani, and Y.~Sato, ``Future person localization in
  first-person videos,'' in \emph{CVPR}, 2018.

\bibitem{yao2018egocentric}
Y.~Yao, M.~Xu, C.~Choi, D.~J. Crandall, E.~M. Atkins, and B.~Dariush,
  ``Egocentric vision-based future vehicle localization for intelligent driving
  assistance systems,'' \emph{ICRA}, 2019.

\bibitem{chandola2009anomaly}
V.~Chandola, A.~Banerjee, and V.~Kumar, ``Anomaly detection: A survey,'' in
  \emph{ACM CSUR}, 2009.

\bibitem{hasan2016learning}
M.~Hasan, J.~Choi, J.~Neumann, A.~K. Roy-Chowdhury, and L.~S. Davis, ``Learning
  temporal regularity in video sequences,'' in \emph{CVPR}, 2016.

\bibitem{medel2016anomaly}
J.~R. Medel and A.~Savakis, ``Anomaly detection in video using predictive
  convolutional long short-term memory networks,'' \emph{arXiv:1612.00390},
  2016.

\bibitem{chong2017abnormal}
Y.~S. Chong and Y.~H. Tay, ``Abnormal event detection in videos using
  spatiotemporal autoencoder,'' in \emph{ISNN}, 2017.

\bibitem{luo2017revisit}
W.~Luo, W.~Liu, and S.~Gao, ``A revisit of sparse coding based anomaly
  detection in stacked rnn framework,'' in \emph{ICCV}, 2017.

\bibitem{herzig2018classifying}
R.~Herzig, E.~Levi, H.~Xu, E.~Brosh, A.~Globerson, and T.~Darrell,
  ``Classifying collisions with spatio-temporal action graph networks,''
  \emph{arXiv:1812.01233}, 2018.

\bibitem{xu2018temporal}
M.~Xu, M.~Gao, Y.-T. Chen, L.~S. Davis, and D.~J. Crandall, ``Temporal
  recurrent networks for online action detection,'' \emph{ICCV}, 2018.

\bibitem{wojke2017simple}
N.~Wojke, A.~Bewley, and D.~Paulus, ``Simple online and realtime tracking with
  a deep association metric,'' in \emph{ICIP}, 2017.

\bibitem{li2014anomaly}
W.~Li, V.~Mahadevan, and N.~Vasconcelos, ``Anomaly detection and localization
  in crowded scenes,'' \emph{TPAMI}, 2014.

\bibitem{lu2013abnormal}
C.~Lu, J.~Shi, and J.~Jia, ``Abnormal event detection at 150 fps in matlab,''
  in \emph{ICCV}, 2013.

\bibitem{Sultani_2018_CVPR}
W.~Sultani, C.~Chen, and M.~Shah, ``Real-world anomaly detection in
  surveillance videos,'' in \emph{CVPR}, 2018.

\bibitem{ananxingche}
\url{https://www.youtube.com/channel/UC-Oa3wml6F3YcptlFwaLgDA/featured}.

\bibitem{he2017mask}
K.~He, G.~Gkioxari, P.~Doll{\'a}r, and R.~Girshick, ``{Mask R-CNN},'' in
  \emph{ICCV}, 2017.

\bibitem{pytorch}
\url{http://pytorch.org/}.

\bibitem{mur2017orb}
R.~Mur-Artal and J.~D. Tard{\'o}s, ``Orb-slam2: An open-source slam system for
  monocular, stereo, and rgb-d cameras,'' \emph{T-RO}, 2017.

\bibitem{ilg2017flownet}
E.~Ilg, N.~Mayer, T.~Saikia, M.~Keuper, A.~Dosovitskiy, and T.~Brox, ``Flownet
  2.0: Evolution of optical flow estimation with deep networks,'' in
  \emph{CVPR}, 2017.

\bibitem{chung2015gated}
J.~Chung, C.~Gulcehre, K.~Cho, and Y.~Bengio, ``Gated feedback recurrent neural
  networks,'' in \emph{ICML}, 2015.

\bibitem{hinton2012neural}
G.~Hinton, N.~Srivastava, and K.~Swersky, ``Neural networks for machine
  learning, lecture 6a: Overview of mini--batch gradient descent.''

\bibitem{carreira2017quo}
J.~Carreira and A.~Zisserman, ``Quo vadis, action recognition? a new model and
  the kinetics dataset,'' in \emph{CVPR}, 2017.

\end{thebibliography}

\end{document}